\pdfoutput=1
\documentclass[11pt]{article}

\usepackage{emnlp2021}
\usepackage{times}
\usepackage{latexsym}
\usepackage{graphicx}
\usepackage{footnote}
\usepackage{multirow}
\usepackage{comment}

\usepackage[T1]{fontenc}
\usepackage[utf8]{inputenc}

\usepackage{microtype}

\usepackage{xspace}

\newcommand{\ours}{\textsc{Evoquer}\xspace}
\newcommand{\rui}[1]{\textcolor{blue}{\emph{[Rui: #1]}}}
\newcommand{\yanjun}[1]{\textcolor{purple}{\emph{[Yanjun: #1]}}}

\title{\ours: Enhancing Temporal Grounding with Video-Pivoted Back Query Generation}

\author{Yanjun Gao$^{1}$, Lulu Liu$^{1}$, Jason Wang$^{1}$, Xin Chen$^{2}$, Huayan Wang$^{2}$, Rui Zhang$^{1}$ \\
        Pennsylvania State University$^{1}$, Kwai Inc$^{2}$ \\ 
        \tt { \{yug125,lzl5409,jjw6188,rmz5227\}@psu.edu},\\
        \tt {xinchen.hawaii@gmail.com, wanghuayan@kuaishou.com}}

\begin{document}
\maketitle
\begin{abstract}
Temporal grounding aims to predict a time interval of a video clip corresponding to a natural language query input. In this work, we present \ours, a temporal grounding framework incorporating an existing text-to-video grounding model and a video-assisted query generation network. Given a query and an untrimmed video, the temporal grounding model predicts the target interval, and the predicted video clip is fed into a video translation task by generating a simplified version of the input query. 
% Our framework forms closed-loop learning by taking into account both the loss functions from the temporal grounding task and loss functions from the translation, serving as feedback. 
\ours forms closed-loop learning by incorporating loss functions from both temporal grounding and query generation serving as feedback.
Our experiments on two widely used datasets, Charades-STA and ActivityNet, show that \ours achieves promising improvements by 1.05 and 1.31 at R@0.7. We also discuss how the query generation task could facilitate error analysis by explaining temporal grounding model behavior.
% and limitations of current architecture.   
%\rui{include some numbers}
% \rui{emphasize model interpretability benefit; also mention activitynet}
% We also present error analysis and discuss the future direction. 
\end{abstract}

\section{Introduction}
%\yanjun{Need to emphasize that our work is based on a simple assumption: two tasks could form a closed-loop learning by integrating the input/output flow and combining the loss; it is a simple architecture, but it works on both datasets; and probably need to have some related work on meta-learning; multi-task learning}
Temporal grounding aims to find the time interval in an untrimmed video that expresses the same meaning as a natural language query. It locates the video content that semantically corresponds to a natural language query, addressing the temporal, semantic alignment between language and vision. It is broadly applicable in many tasks such as visual storytelling~\cite{lukin-etal-2018-pipeline,huang2016visual}, video caption generation~\cite{krishna2017dense,long2018video}, and video machine translation~\cite{wang2019vatex}.

Recent work on temporal grounding has achieved significant progress~\cite{mun2020local,chen2019semantic,chen2018temporally,zhang2019man,gao2017tall}.
They emphasize modeling the semantic mapping of verbs and nouns in the text query to visual clues such as actions and objects that indicate the candidate time intervals. However, most of them only employ a uni-direction single-task learning flow. %Inspired by multi-task learning for temporal grounding~\cite{xu2019multilevel}\rui{differentiate ours from this work}, 
% \textcolor{red}{
To strengthen the learning and facilitate error analysis, we explore the possibility of enhancing the temporal grounding model with related tasks.
To this end, we borrow the idea of feedback-error-learning from control theory and computational neuroscience~\cite{kawato1990feedback,gomi1993neural}. Using a closed-loop system, the control network learns to correct its error from feedback and gains stronger supervision to stimulate learning.  
We investigate if a temporal grounding model can be improved by incorporating another network to generate feedback in a closed-loop learning fashion.

\begin{figure}[t!]
\centering
%\begin{center}
%\vspace{-.1in}
\includegraphics[scale=0.4]{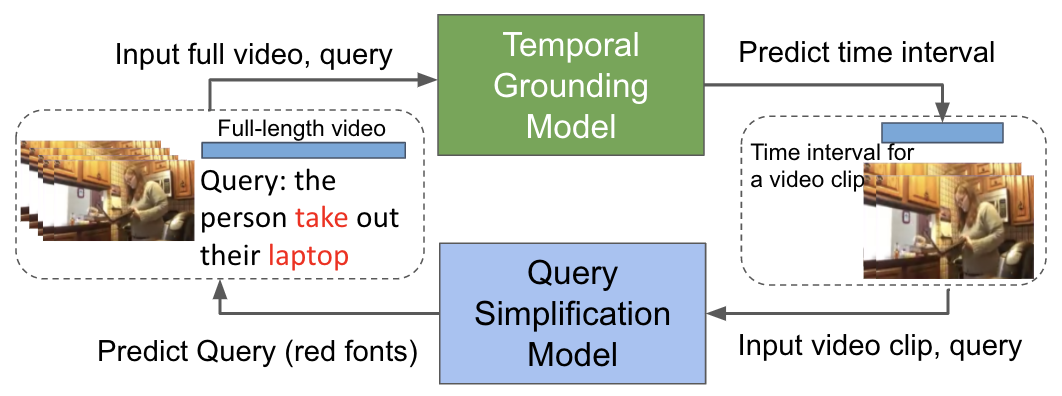} 
% \vspace{-1pc}
\vspace{-0.7pc}
\caption{An overview of \ours as a closed-loop system pipeline.} 
% \vspace{-2pc}
\label{fig:pipeline}
\end{figure}

\begin{figure*}[t!]
\centering
\includegraphics[scale=0.38]{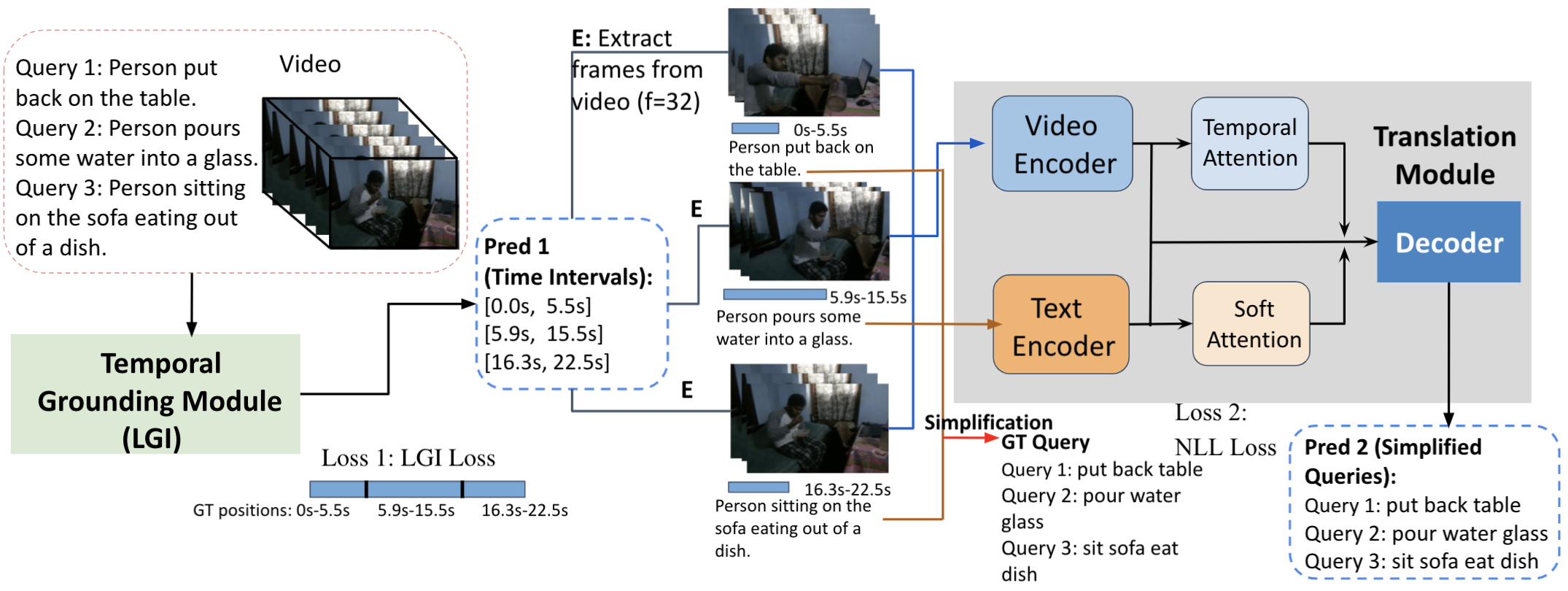} 
\vspace{-0.7pc}
\caption{\small Our \ours framework combines a LGI model for temporal grounding and a translation module that outputs simplified queries.} 
\label{fig:architecture}
\end{figure*}

% Chinese Whisper
We propose a novel framework, \ours (\textit{E}nhancing Temporal Grounding with \textit{V}ide\textit{O}-Pivoted Back \textit{QUER}y Generation), integrating a text-to-video and a video-to-text flow, as shown in Figure~\ref{fig:pipeline}. Specifically, we adapt a video-pivoted query simplification task that simplifies the query to shorter phrases with verbs and noun phrases only. 
% \textcolor{red}{
Instead of re-generating the full queries, the query simplification task is smaller in the problem size and thus could serve as an auxiliary task to assist the main task. Furthermore, we incorporate visual pivots in query simplification to provide more fine-grained semantic discrepancy associated with words~\cite{chen2019words,lee2019countering}. The pipeline pairs a state-of-the-art temporal grounding model LGI~\cite{mun2020local} with a video machine translation model~\cite{wang2019vatex} for query simplification. Given a query and an untrimmed video, the pipeline predicts the time interval, feeds the predicted video clips with the original query to the translation model, and generates a simplified query. 
% \textcolor{red}{
\ours generates two losses individually from the temporal grounding and query simplification task, and combines the loss to update all network components, giving stronger supervision signals. On two temporal grounding datasets, Charades-STA~\cite{gao2017tall} and ActivityNet~\cite{krishna2017dense}, \ours outperforms the original model by 1.05 and 1.31 at R@0.7, demonstrating its effectiveness. Our analysis of query simplification output indicates that this video-to-text auxiliary task casts light on explaining temporal grounding model behavior.
% }
%\rui{include some numbers}
%We evaluate \ours on two temporal grounding datasets Charades-STA and ActivityNet, demonstrate promising results from interval prediction performance and qualities of simplification output, and analyze the output for future improvements.
%\rui{mention a bit about experiments}
\section{Related Work}
Existing temporal grounding approaches can be split into three categories: strongly supervised ~\cite{anne2017localizing,gao2017tall,liu2018attentive,chen2018temporally,chen2019localizing,chen2019semantic,ge2019mac,ghosh2019excl,zhang2019man,yuan2019find,mun2020local,rodriguez2020proposal}, weakly supervised~\cite{tan2021logan}, and reinforcement learning~\cite{wang2019language,he2019read}. 
% Our work belongs to the supervised learning framework. %The Latent Graph Co-Attention Network (LoGAN) approach utilizes a weakly supervised approach by modeling context between pairs of video frames~\cite{tan2021logan}.% Although not associated with our the \ours framework, other approaches have utilized a reinforcement learning based framework with multi-task learning. These approaches have demonstrated steady performance improvements~\cite{he2019read}. 
The primary example of strongly supervised approaches is LGI~\cite{mun2020local} that achieves state-of-the-art performance on the Charades-STA and ActivityNet datasets.
It uses word-level and sentence-level attention to predict time intervals.
Within our closed-loop framework, we utilize this LGI algorithm as a black box to achieve the best performance on supervised temporal grounding. %LGI algorithm is mostly treated as a "black box".
%\paragraph{Text-to-Video Retrieval} 
Recent progress also uses transformers with language and vision pretraining~\cite{radford2021learning,lei2021less,luo2021clip4clip}.

Another similar task is text-to-video moment retrieval that focuses on grounding between query and video~\cite{xu2019multilevel,lin2020weakly,liu2018attentive}, framing the task as retrieving video frames, slightly different from temporal grounding.  
%Text-to-Clip retrieval methods have also been provided by other strongly supervised approaches. Within this approach, a segment proposal network is used to filter out unlikely and redundant clips.
%\paragraph{Video Captioning}
Furthermore, our framework is also relevant to video captioning which aims to generate a description of text given a video~\cite{das2013thousand,yao2015describing,venugopalan2015sequence,venugopalan2015translating,xu2015multi,zhou2019grounded,zhou2018towards}. Recent developments have utilized end-to-end transformer models for video captioning~\cite{zhou2018end}. %Too old: ~\citet{donahue2015long} utilized "doubly deep" models in that the approach learns representations in space and time. 
\section{\ours Framework} 
We design a closed-loop framework for temporal grounding such that the model receives (1) supervision in predicting time intervals and (2) feedback from the output video features extracted from the prediction.
To achieve this, \ours involves two components: a temporal grounding module and a translation module. The temporal grounding module predicts time intervals given an untrimmed video and a query. The translation module takes input from queries and video features trimmed by the predicted intervals, and outputs a simplified query with only verbs and nouns. 
We use the LGI model~\cite{mun2020local} for temporal grounding, which achieves state-of-the-art performance using supervised learning. For query simplification, we use the video machine translation framework VMT~\cite{wang2019vatex} whose performance is competitive in video-assisted bilingual translation. 
%VMT is proposed for video-assisted bilingual translation, for example, between Chinese and English, and achieves competitive results.

Our pipeline is presented in Figure~\ref{fig:architecture}. The input to the framework is an untrimmed video and a set of queries. Following~\citet{mun2020local}, we use I3D frame-based features for video representation and an embedding layer inside a text encoder for word representation. Given the video features and queries, LGI predicts time intervals with the content corresponding to a given query. Next, we extract frames from videos trimmed by the predicted interval to represent the content of the video clip. To maintain the continuity of the content, we extract 32 frames per video clip in a way that the content of the trimmed videos is evenly distributed across all 32 frames. Since the camera used captures 24 frames per second, a 32-frame video roughly spans 1.3 seconds. We feed the extracted video features and input query into a translation module consisting of two biLSTM-based encoders and an LSTM-based decoder with attention. Video hidden states and text hidden states are sent individually to two attention modules, while being concatenated into one vector representation and sent to the decoder as initial hidden states. In the attention network, temporal attention is learned through video features, and soft attention through query hidden states. The attention is fed into the decoder as context representation.

Instead of learning to decode the original query, we want the model to focus on the words that distinguish the video content: verbs and nouns. In the Charades dataset, annotators tend to use various verb tenses when describing the video activities. For example, both \textit{``closes the door"} and \textit{``closing the door"} are used on the same video content. Therefore, we lemmatize the words, label the query with part-of-speech (POS) tags, and extract verbs and nouns as simplified versions of the queries. The decoder learns to predict simplified queries and computes a negative log-likelihood (NLL) loss at the end of the decoding. Finally, we combine the NLL loss from query simplification with the LGI loss from time interval prediction to train the networks jointly in an end-to-end fashion.
% computed earlier to update the networks.

%In addition, we experiment with an alternative setting of the translation module: we only generate simplified queries from the video input and add a loss to explicitly enforce the mapping between video features and text features. We keep the text encoder to generate text hidden states, and apply a visual embedding (VSE) loss proposed in~\citet{faghri2018vse++} to learn the joint embedding between text and video based on their cosine similarity. The neural network is trained end-to-end to jointly optimize the three loss functions (LGI, NLL, VSE).
% This loss is added to the two losses and propagated to every parameter of the network. 
\section{Experiments and Results}
% \rui{include activitynet / remove VSE}
% In this section, we present our experiments on both Charades-STA \textcolor{red}{and ActivityNet} datasets using \ours. %and \ours+VSE. 
% Results include time interval prediction and simplification output. \ours shows promising improvements over the other settings. %This analysis shows the benefits of having an integrated framework and identifies a few problems for future improvements. 

\paragraph{Datasets}
We evaluate our framework on Charades-STA~\cite{gao2017tall} and Activity Net~\cite{krishna2017dense}, two widely used benchmark data sets for temporal grounding. %Charades-STA is comprised of 9,848 roughly 30-second videos of daily human activities.
%Each video in Charades-STA corresponds to a set of queries created by annotators watching these videos, with each having a maximum length of 10 words. 
%There are 27,847 textual queries in total provided for the videos, with each having a maximum length of 10 words.
%ActivityNet contains 20k videos with average length as 120 seconds. Each video contains roughly 3.7 time intervals with 13.5-words query per interval. 
We follow the dataset setting in~\cite{mun2020local}, where both datasets are set with train/valid/test as 50\%, 25\%, and 25\% respectively. The dataset statistics are reported in Table~\ref{tab:datasetinfo}.  

The two datasets vary greatly on most of the statistics. 
%Charades-STA has 560 words in the simplified vocabulary, with average length as 2.31 tokens per query. For ActivityNet, the simplified query vocabulary contains 5,946 words, and the average length of simplified query is 4.12 tokens. 
We think ActivityNet is a more challenging dataset, as it requires the decoder in \ours to predict correct words from a much bigger vocabulary whose size is almost 10 times bigger than Charades-STA. In our experiments on ActivityNet, we find that the \ours converges at a higher NLL loss than Charades-STA and it fails to produce good quality simplified queries, which we suspect is attributed to the harder decoding task. %We will show in the next section that it is harder to train \ours on ActivityNet and obtain good quality output of query simplification. % It is hard for auxiliary task to achieve good performance when the problem size is big.

%\rui{Use a table to summarize the stats of two datasets?}
%\yanjun{There was one table; taking too much room}
\begin{comment}
\rui{Train/Test split? Number of queries per video? What's the difference between ``textual descriptions" and ``text queries"? Are we using both or just ``text queries"?}
\yanjun{Jason/Lulu could fill in the stats}
\end{comment}

\begin{table}[t]
\small
    \centering
    \begin{tabular}{l|rr} \hline 
    Dataset & Charades-STA & ActivityNet \\ \hline
    Num Queries & 27,847 & 71,957 \\
    Num Videos & 9,848 & 20,000 \\
    Avg Video Len (Sec) & \textasciitilde 30  & \textasciitilde 120\\
    Input Query |V| & 1,140 & 11,125\\
    Simpl. Query |V| & 560 & 5,946\\
    Simpl. \#Tks per Query & 2.31 & 4.12\\ \hline
    \end{tabular}
    \vspace{-0.8pc}
    \caption{\small Statistics contrasting Charades-STA and ActivityNet and the impact of simplification on each dataset.}
    \label{tab:datasetinfo}
\end{table}

% \begin{table}[t]
% \small
%     \centering
%     \begin{tabular}{l|rrrrr} \hline 
%     Data& Query |V| & Simpl. |V|& Query\#Tks & Simpl.\#Tks  \\ \hline
%     char & 1,140 & 560 &  & 2.31\\
%     anet & 11,125 & 5,946 & & 4.12\\ \hline 
%     \end{tabular}
%     \vspace{-0.8pc}
%     \caption{The impact of simplification on Charades-STA (char) and ActivityNet (anet) vocabulary size.}
%     \label{tab:datasetinfo}
% \end{table}

\paragraph{Evaluation metrics}
We adopt two conventional temporal grounding metrics: \textit{R@tIoU} measuring recall at different thresholds (0.3, 0.5, and 0.7) for temporal intervals between ground truth and prediction; \textit{mIoU} reporting the average of temporal interval recall from all threshold levels. For query simplification, we evaluate the predicted queries with two metrics. Jaccard similarity measures intersection over union between words in ground truth and in prediction. Since it does not penalize for duplicated words, Jaccard similarity gives us a rough estimation for the quality of translation output. BLEU~\cite{papineni2002bleu} is a standard evaluation metric for machine translation that measures n-gram word overlap. Most of the simplified queries are two-word length, thus we report BLEU unigram and bigram. 

% We train our model with the same set of parameters presented in~\citet{mun2020local}.  

\paragraph{Temporal grounding results}

Table~\ref{tab:results} presents results on the Charades-STA and ActivityNet test set from a re-trained LGI model and \ours models.\footnote{Using the codes from the author's GitHub and the parameters presented in the original paper, we train the LGI model on Charades-STA train set from scratch. We suspect the difference between our replication and results presented in the paper is attributed to initialization.} Compared to LGI, \ours shows improvement on R@0.7 and mIoU, especially 1.05 and 1.31 on R@0.7, the hardest threshold for temporal interval overlap.%\footnote{Parameter setting is reported in Appendix.} %\ours also outperforms LGI on R@0.3 and mIoU. %however , there are some drops on R@0.3 and mIoU with VSE.    

\begin{table}[t]
\small
    \centering
    \begin{tabular}{ll|rrrr} \hline 
    Data& Model & R@0.3 & R@0.5 & R@0.7 & mIoU \\ \hline 
    \multirow{2}{*}{char} & LGI & 71.54 & \textbf{58.08} & 34.68 & 50.28 \\ 
    & \ours & \textbf{71.57}& 57.81& \textbf{35.73} & \textbf{50.48} \\ 
    \multirow{2}{*}{anet} & LGI & 57.76 & 40.38 & 22.60 & 40.65 \\ 
    & \ours & \textbf{59.21} & \textbf{42.02} & \textbf{23.91} & \textbf{41.61} \\ \hline 
    %\ours+VSE & 70.46 & 57.81 & 35.51 &50.16 \\ \hline 
    \end{tabular}
    \vspace{-0.8pc}
    \caption{\small Results on Charades-STA (char) and ActivityNet (anet) from the LGI model and \ours. }
    \label{tab:results}
\end{table}

\begin{table}[t]
\small
    \centering
    \begin{tabular}{l|rrrrr} \hline 
    & \multicolumn{3}{c}{Both $>=$ R@0.3 } & Both \\ 
    & \ours$\uparrow$ & \ours$\downarrow$ & Same & $<$R@0.3 \\ \hline 
    char & 441 & 362 & 1,347 & 777 \\ 
    anet & 4,268 & 3,124 & 8,074 & 10,538 \\ 
    \hline
    \end{tabular}
    \vspace{-0.5pc}
    \caption{\small Counts of samples that are scored by R@tIoU with four categories from comparison between \ours and LGI model. Three of the categories are from samples where both models achieve recall $>=$ threshold 0.3: samples that are improved (\ours$\uparrow$), samples with performance drops (\ours$\downarrow$), and equal performance with at least R@0.3 (Same). The fourth category is when both perform below R@0.3 (Both <R@0.3).}
    \label{tab:types}
\end{table}

Table~\ref{tab:types} presents statistics of samples where our model show improvements and drops compared to LGI. We divide the samples into four categories according to their recall: when \ours ranks in a higher threshold than the LGI (e.g. 0.7 vs 0.5), when \ours ranks lower than the LGI, when both have the same recalls that are at least R@0.3, and when both scores rank below R@0.3. \ours shows 441 samples on Charades-STA and 4,268 samples of ActivityNet that are above 0.3 recall threshold, with 79 and 1,144 samples of absolute improvement. There remains a large number of samples to be improved (Both <R@0.3)). Recall that query simplification task serves as an auxiliary task for the temporal grounding model thus the performance upper bound of \ours could be limited by LGI. We consider the \ours improvements promising after seeing a fair number of samples being improved.  %Recall that \ours performance is still limited by LGI performance and thus there remains a large number of samples to be improved (Both <R@0.3)). %This preliminary results show that \ours could bring potential benefits to the temporal grounding task. %There are 777 cases where both models perform poorly, showing more room for improvements. To summarize, this preliminary results show that \ours could bring potential benefits to the temporal grounding task.       

\paragraph{Translation output analysis}

% \begin{table}[t]
% \small
%     \centering
%     \begin{tabular}{l|rrr} \hline 
%     Model & JaccSim & BLEU1 & BLEU2  \\ \hline 
%     \ours & \textbf{51.98}& \textbf{53.04} & \textbf{42.47} \\
%     \ours+VSE & 6.37 & 7.96 & 1.20  \\ \hline 
%     \end{tabular}
%     \vspace{-0.5pc}
%     \caption{\small Translation quality measuring by Jaccard similarity, BLEU Unigram (BLEU1) and Bigram (BLEU2).}
%     \label{tab:bleu}
% \end{table}

%We compare translation output from  \ours variants. This comparison could help us identify the components that are critical to video-assisted machine translation. 
% We evaluate the quality of translation output using BLEU and Jaccard scores, both of which measure the word overlap between ground truth and prediction. 
We evaluate translation outputs using BLEU and Jaccard scores between ground truths and predictions.
On Charades-STA test set, \ours achieved 51.98 Jaccard Similarity, 53.04 on BLEU unigram, and 42.47 on BLEU bigram. We will show how the query simplification output helps the error analysis of temporal grounding model in case study. On ActivityNet, we fail at training \ours to generate reasonable simplified queries. We suspect two reasons for this failure: 1) the decoder vocabulary on ActivityNet is much larger than Charades-STA, which makes the simplification task harder, as we mentioned previously; 2) the current design of translation module is too simple to handle features from the longer  predicted time interval and the input queries. Nonetheless, the results on Charades-STA indicates the potential benefit of our framework, shown in case studies next. %We assume that using pre-trained knowledge for  
%due to the large size of decoder vocabulary. 
%Results are shown in Table~\ref{tab:bleu}. Although both frameworks show similar trends in performance of the temporal grounding task, their translation quality have a large difference:\ours shows good scores, while \ours+VSE shows significantly lower performance. This shows that both video features and text features are critical to translation.   
\paragraph{Case study}
% \rui{talk less about VSE; more about interpretability benefit}
\begin{figure}[t!]
\centering
%\begin{center}
%\vspace{-.1in}
\includegraphics[width=\columnwidth]{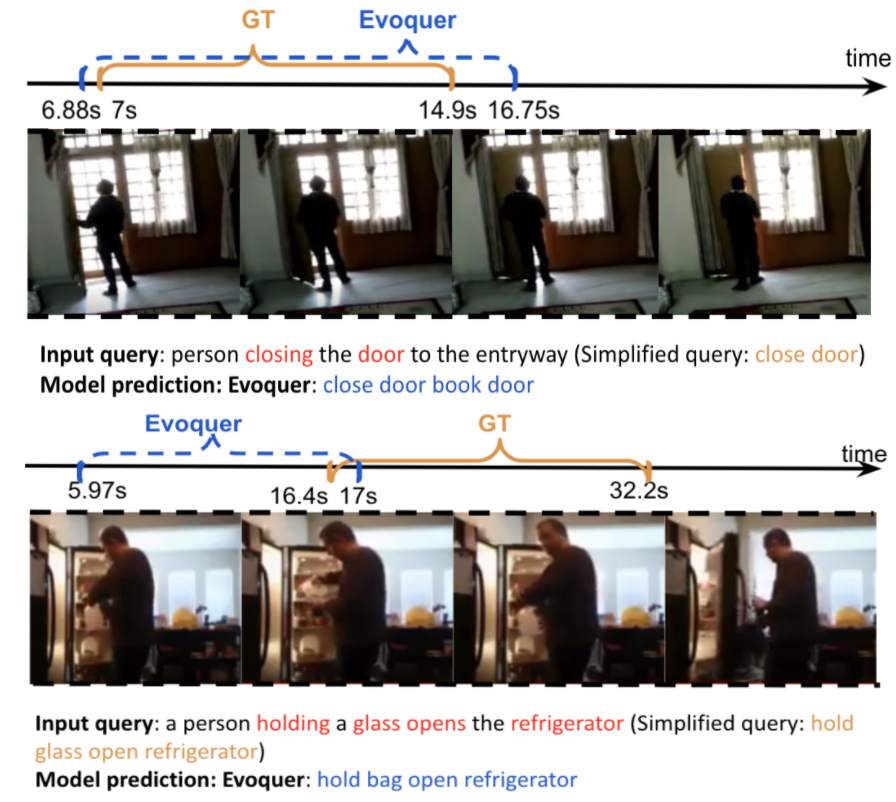} 
\vspace{-2pc}
\caption{\small Two example video clips trimmed by ground truth intervals. In the first example (top), \ours successfully predicts time interval and simplified queries as ground truth. In the second example (bottom), \ours fails to predict time interval. %Simplified queries predicted by \ours+VSE are also presented.  
} 
\label{fig:example}
\end{figure}

% \begin{figure}[t!]
% \centering
% %\begin{center}
% %\vspace{-.1in}
% \includegraphics[width=\columnwidth]{failure.png} 
% \vspace{-2pc}
% \caption{\small Another example of video clips trimmed by ground truth interval. \ours fails to predict time interval.} 
% \label{fig:good_example}
% \end{figure}

% In this section, w
We show output examples of predicted intervals and simplified queries to understand the model performance. Figure~\ref{fig:example} shows two video clips trimmed by the ground truth interval, the queries, and predicted simplification. In the first example, \ours predicts interval overlapping with ground truth and correctly translates the verb and noun \textit{close door}. %\ours+VSE inaccurately predicts the verb \textit{open} instead of \textit{close}. 
Judging from the video content, the door was already closed; thus, an \textit{open door} action must occur before the \textit{close door}. %Given that \ours+VSE only takes video content as input to decoder, we suspect the features of \textit{open door} are stronger than \textit{close door} thus captured by the decoder in \ours+VSE. 
In the second case, \ours predicts an interval rarely intersecting with ground truth. We review the video and find that at 5.97s, the person in the video starts the action \textit{open the refrigerator door} and pours milk into a glass. Additionally, at 16.4s, he finishes \textit{pouring} and puts the milk back into the refrigerator (shown as the first picture of Figure 3 bottom). Meanwhile, he is holding the glass and leaving the refrigerator door open. Although \ours fails to intersect with the gold standard, it captures the action \textit{open the door} at 5.97s, showing its capability in understanding the video content. We suspect \ours thinks that the person is holding a bag instead of a gallon of milk since both are white in color and similar in size. Thus, it predicts \textit{hold bag} instead of \textit{hold glass}. 
% Our future work will extend the experiments on other temporal grounding datasets to better validate \ours performance.

% \begin{table}[t]
% \small
%     \centering
%     \begin{tabular}{l|lll} \hline 
%     (QID) GT & Model & Pred   \\ \hline 
%     %Ours & 2633 & close door & close cabinet take door \\ 
%      (2007) start camera & Ours & start camera play read\\
%       play& Ours+VSE & put picture put shoe \\
%      \multirow{2}{*}{(2619) turn light} & Ours & turn light door take\\
%     & Ours+VSE &  open door open door \\ \hline 
%     \end{tabular}
%     \vspace{-0.5pc}
%     \caption{Example of query output from two models}
%     \label{tab:bleu}
% \end{table}

\section{Conclusion}
We propose a novel framework, \ours, for temporal grounding that incorporates a query simplification task. It forms closed-loop learning and provides feedback to the temporal grounding model and enhances the learning. Our experiments demonstrate promising results on predicting time intervals and query simplification. Future work will explore more settings and extend to other datasets.

% Entries for the entire Anthology, followed by custom entries
\bibliography{anthology,main}

\begin{thebibliography}{38}
\expandafter\ifx\csname natexlab\endcsname\relax\def\natexlab#1{#1}\fi

\bibitem[{Anne~Hendricks et~al.(2017)Anne~Hendricks, Wang, Shechtman, Sivic,
  Darrell, and Russell}]{anne2017localizing}
Lisa Anne~Hendricks, Oliver Wang, Eli Shechtman, Josef Sivic, Trevor Darrell,
  and Bryan Russell. 2017.
\newblock Localizing moments in video with natural language.
\newblock In \emph{Proceedings of the IEEE international conference on computer
  vision}, pages 5803--5812.

\bibitem[{Chen et~al.(2018)Chen, Chen, Ma, Jie, and Chua}]{chen2018temporally}
Jingyuan Chen, Xinpeng Chen, Lin Ma, Zequn Jie, and Tat-Seng Chua. 2018.
\newblock Temporally grounding natural sentence in video.
\newblock In \emph{Proceedings of the 2018 conference on empirical methods in
  natural language processing}, pages 162--171.

\bibitem[{Chen et~al.(2019{\natexlab{a}})Chen, Ma, Chen, Jie, and
  Luo}]{chen2019localizing}
Jingyuan Chen, Lin Ma, Xinpeng Chen, Zequn Jie, and Jiebo Luo.
  2019{\natexlab{a}}.
\newblock Localizing natural language in videos.
\newblock In \emph{Proceedings of the AAAI Conference on Artificial
  Intelligence}.

\bibitem[{Chen and Jiang(2019)}]{chen2019semantic}
Shaoxiang Chen and Yu-Gang Jiang. 2019.
\newblock Semantic proposal for activity localization in videos via sentence
  query.
\newblock In \emph{Proceedings of the AAAI Conference on Artificial
  Intelligence}.

\bibitem[{Chen et~al.(2019{\natexlab{b}})Chen, Jin, and Fu}]{chen2019words}
Shizhe Chen, Qin Jin, and Jianlong Fu. 2019{\natexlab{b}}.
\newblock From words to sentences: A progressive learning approach for
  zero-resource machine translation with visual pivots.
\newblock \emph{arXiv preprint arXiv:1906.00872}.

\bibitem[{Das et~al.(2013)Das, Xu, Doell, and Corso}]{das2013thousand}
Pradipto Das, Chenliang Xu, Richard~F Doell, and Jason~J Corso. 2013.
\newblock A thousand frames in just a few words: Lingual description of videos
  through latent topics and sparse object stitching.
\newblock In \emph{Proceedings of the IEEE conference on computer vision and
  pattern recognition}, pages 2634--2641.

\bibitem[{Gao et~al.(2017)Gao, Sun, Yang, and Nevatia}]{gao2017tall}
Jiyang Gao, Chen Sun, Zhenheng Yang, and Ram Nevatia. 2017.
\newblock Tall: Temporal activity localization via language query.
\newblock In \emph{Proceedings of the IEEE international conference on computer
  vision}, pages 5267--5275.

\bibitem[{Ge et~al.(2019)Ge, Gao, Chen, and Nevatia}]{ge2019mac}
Runzhou Ge, Jiyang Gao, Kan Chen, and Ram Nevatia. 2019.
\newblock Mac: Mining activity concepts for language-based temporal
  localization.
\newblock In \emph{2019 IEEE Winter Conference on Applications of Computer
  Vision (WACV)}, pages 245--253. IEEE.

\bibitem[{Ghosh et~al.(2019)Ghosh, Agarwal, Parekh, and
  Hauptmann}]{ghosh2019excl}
Soham Ghosh, Anuva Agarwal, Zarana Parekh, and Alexander Hauptmann. 2019.
\newblock Excl: Extractive clip localization using natural language
  descriptions.
\newblock \emph{arXiv preprint arXiv:1904.02755}.

\bibitem[{Gomi and Kawato(1993)}]{gomi1993neural}
Hiroaki Gomi and Mitsuo Kawato. 1993.
\newblock Neural network control for a closed-loop system using
  feedback-error-learning.
\newblock \emph{Neural Networks}, 6(7):933--946.

\bibitem[{He et~al.(2019)He, Zhao, Huang, Li, Liu, and Wen}]{he2019read}
Dongliang He, Xiang Zhao, Jizhou Huang, Fu~Li, Xiao Liu, and Shilei Wen. 2019.
\newblock Read, watch, and move: Reinforcement learning for temporally
  grounding natural language descriptions in videos.
\newblock In \emph{Proceedings of the AAAI Conference on Artificial
  Intelligence}.

\bibitem[{Huang et~al.(2016)Huang, Ferraro, Mostafazadeh, Misra, Agrawal,
  Devlin, Girshick, He, Kohli, Batra et~al.}]{huang2016visual}
Ting-Hao Huang, Francis Ferraro, Nasrin Mostafazadeh, Ishan Misra, Aishwarya
  Agrawal, Jacob Devlin, Ross Girshick, Xiaodong He, Pushmeet Kohli, Dhruv
  Batra, et~al. 2016.
\newblock Visual storytelling.
\newblock In \emph{Proceedings of the 2016 Conference of the North American
  Chapter of the Association for Computational Linguistics: Human Language
  Technologies}, pages 1233--1239.

\bibitem[{Kawato(1990)}]{kawato1990feedback}
Mitsuo Kawato. 1990.
\newblock Feedback-error-learning neural network for supervised motor learning.
\newblock In \emph{Advanced neural computers}, pages 365--372. Elsevier.

\bibitem[{Krishna et~al.(2017)Krishna, Hata, Ren, Fei-Fei, and
  Niebles}]{krishna2017dense}
Ranjay Krishna, Kenji Hata, Frederic Ren, Li~Fei-Fei, and Juan~Carlos Niebles.
  2017.
\newblock Dense-captioning events in videos.
\newblock In \emph{International Conference on Computer Vision (ICCV)}.

\bibitem[{Lee et~al.(2019)Lee, Cho, and Kiela}]{lee2019countering}
Jason Lee, Kyunghyun Cho, and Douwe Kiela. 2019.
\newblock Countering language drift via visual grounding.
\newblock In \emph{Proceedings of the 2019 Conference on Empirical Methods in
  Natural Language Processing and the 9th International Joint Conference on
  Natural Language Processing (EMNLP-IJCNLP)}, pages 4376--4386.

\bibitem[{Lei et~al.(2021)Lei, Li, Zhou, Gan, Berg, Bansal, and
  Liu}]{lei2021less}
Jie Lei, Linjie Li, Luowei Zhou, Zhe Gan, Tamara~L Berg, Mohit Bansal, and
  Jingjing Liu. 2021.
\newblock Less is more: Clipbert for video-and-language learning via sparse
  sampling.
\newblock \emph{arXiv preprint arXiv:2102.06183}.

\bibitem[{Lin et~al.(2020)Lin, Zhao, Zhang, Wang, and Liu}]{lin2020weakly}
Zhijie Lin, Zhou Zhao, Zhu Zhang, Qi~Wang, and Huasheng Liu. 2020.
\newblock Weakly-supervised video moment retrieval via semantic completion
  network.
\newblock In \emph{Proceedings of the AAAI Conference on Artificial
  Intelligence}.

\bibitem[{Liu et~al.(2018)Liu, Wang, Nie, He, Chen, and
  Chua}]{liu2018attentive}
Meng Liu, Xiang Wang, Liqiang Nie, Xiangnan He, Baoquan Chen, and Tat-Seng
  Chua. 2018.
\newblock Attentive moment retrieval in videos.
\newblock In \emph{The 41st international ACM SIGIR conference on research \&
  development in information retrieval}, pages 15--24.

\bibitem[{Long et~al.(2018)Long, Gan, and De~Melo}]{long2018video}
Xiang Long, Chuang Gan, and Gerard De~Melo. 2018.
\newblock Video captioning with multi-faceted attention.
\newblock \emph{Transactions of the Association for Computational Linguistics},
  6:173--184.

\bibitem[{Lukin et~al.(2018)Lukin, Hobbs, and Voss}]{lukin-etal-2018-pipeline}
Stephanie Lukin, Reginald Hobbs, and Clare Voss. 2018.
\newblock \href {https://doi.org/10.18653/v1/W18-1503} {A pipeline for creative
  visual storytelling}.
\newblock In \emph{Proceedings of the First Workshop on Storytelling}, pages
  20--32, New Orleans, Louisiana. Association for Computational Linguistics.

\bibitem[{Luo et~al.(2021)Luo, Ji, Zhong, Chen, Lei, Duan, and
  Li}]{luo2021clip4clip}
Huaishao Luo, Lei Ji, Ming Zhong, Yang Chen, Wen Lei, Nan Duan, and Tianrui Li.
  2021.
\newblock Clip4clip: An empirical study of clip for end to end video clip
  retrieval.
\newblock \emph{arXiv preprint arXiv:2104.08860}.

\bibitem[{Mun et~al.(2020)Mun, Cho, and Han}]{mun2020local}
Jonghwan Mun, Minsu Cho, and Bohyung Han. 2020.
\newblock Local-global video-text interactions for temporal grounding.
\newblock In \emph{Proceedings of the IEEE/CVF Conference on Computer Vision
  and Pattern Recognition}, pages 10810--10819.

\bibitem[{Papineni et~al.(2002)Papineni, Roukos, Ward, and
  Zhu}]{papineni2002bleu}
Kishore Papineni, Salim Roukos, Todd Ward, and Wei-Jing Zhu. 2002.
\newblock Bleu: a method for automatic evaluation of machine translation.
\newblock In \emph{Proceedings of the 40th annual meeting of the Association
  for Computational Linguistics}, pages 311--318.

\bibitem[{Radford et~al.(2021)Radford, Kim, Hallacy, Ramesh, Goh, Agarwal,
  Sastry, Askell, Mishkin, Clark et~al.}]{radford2021learning}
Alec Radford, Jong~Wook Kim, Chris Hallacy, Aditya Ramesh, Gabriel Goh,
  Sandhini Agarwal, Girish Sastry, Amanda Askell, Pamela Mishkin, Jack Clark,
  et~al. 2021.
\newblock Learning transferable visual models from natural language
  supervision.
\newblock \emph{arXiv preprint arXiv:2103.00020}.

\bibitem[{Rodriguez et~al.(2020)Rodriguez, Marrese-Taylor, Saleh, Li, and
  Gould}]{rodriguez2020proposal}
Cristian Rodriguez, Edison Marrese-Taylor, Fatemeh~Sadat Saleh, Hongdong Li,
  and Stephen Gould. 2020.
\newblock Proposal-free temporal moment localization of a natural-language
  query in video using guided attention.
\newblock In \emph{Proceedings of the IEEE/CVF Winter Conference on
  Applications of Computer Vision}, pages 2464--2473.

\bibitem[{Tan et~al.(2021)Tan, Xu, Saenko, and Plummer}]{tan2021logan}
Reuben Tan, Huijuan Xu, Kate Saenko, and Bryan~A Plummer. 2021.
\newblock Logan: Latent graph co-attention network for weakly-supervised video
  moment retrieval.
\newblock In \emph{Proceedings of the IEEE/CVF Winter Conference on
  Applications of Computer Vision}, pages 2083--2092.

\bibitem[{Venugopalan et~al.(2015{\natexlab{a}})Venugopalan, Rohrbach, Donahue,
  Mooney, Darrell, and Saenko}]{venugopalan2015sequence}
Subhashini Venugopalan, Marcus Rohrbach, Jeffrey Donahue, Raymond Mooney,
  Trevor Darrell, and Kate Saenko. 2015{\natexlab{a}}.
\newblock Sequence to sequence-video to text.
\newblock In \emph{Proceedings of the IEEE international conference on computer
  vision}, pages 4534--4542.

\bibitem[{Venugopalan et~al.(2015{\natexlab{b}})Venugopalan, Xu, Donahue,
  Rohrbach, Mooney, and Saenko}]{venugopalan2015translating}
Subhashini Venugopalan, Huijuan Xu, Jeff Donahue, Marcus Rohrbach, Raymond
  Mooney, and Kate Saenko. 2015{\natexlab{b}}.
\newblock \href {https://doi.org/10.3115/v1/N15-1173} {Translating videos to
  natural language using deep recurrent neural networks}.
\newblock In \emph{Proceedings of the 2015 Conference of the North {A}merican
  Chapter of the Association for Computational Linguistics: Human Language
  Technologies}, pages 1494--1504, Denver, Colorado. Association for
  Computational Linguistics.

\bibitem[{Wang et~al.(2019{\natexlab{a}})Wang, Huang, and
  Wang}]{wang2019language}
Weining Wang, Yan Huang, and Liang Wang. 2019{\natexlab{a}}.
\newblock Language-driven temporal activity localization: A semantic matching
  reinforcement learning model.
\newblock In \emph{Proceedings of the IEEE/CVF Conference on Computer Vision
  and Pattern Recognition}, pages 334--343.

\bibitem[{Wang et~al.(2019{\natexlab{b}})Wang, Wu, Chen, Li, Wang, and
  Wang}]{wang2019vatex}
Xin Wang, Jiawei Wu, Junkun Chen, Lei Li, Yuan-Fang Wang, and William~Yang
  Wang. 2019{\natexlab{b}}.
\newblock Vatex: A large-scale, high-quality multilingual dataset for
  video-and-language research.
\newblock In \emph{Proceedings of the IEEE/CVF International Conference on
  Computer Vision}, pages 4581--4591.

\bibitem[{Xu et~al.(2019)Xu, He, Plummer, Sigal, Sclaroff, and
  Saenko}]{xu2019multilevel}
Huijuan Xu, Kun He, Bryan~A Plummer, Leonid Sigal, Stan Sclaroff, and Kate
  Saenko. 2019.
\newblock Multilevel language and vision integration for text-to-clip
  retrieval.
\newblock In \emph{Proceedings of the AAAI Conference on Artificial
  Intelligence}.

\bibitem[{Xu et~al.(2015)Xu, Venugopalan, Ramanishka, Rohrbach, and
  Saenko}]{xu2015multi}
Huijuan Xu, Subhashini Venugopalan, Vasili Ramanishka, Marcus Rohrbach, and
  Kate Saenko. 2015.
\newblock A multi-scale multiple instance video description network.
\newblock \emph{arXiv preprint arXiv:1505.05914}.

\bibitem[{Yao et~al.(2015)Yao, Torabi, Cho, Ballas, Pal, Larochelle, and
  Courville}]{yao2015describing}
Li~Yao, Atousa Torabi, Kyunghyun Cho, Nicolas Ballas, Christopher Pal, Hugo
  Larochelle, and Aaron Courville. 2015.
\newblock Describing videos by exploiting temporal structure.
\newblock In \emph{Proceedings of the IEEE international conference on computer
  vision}, pages 4507--4515.

\bibitem[{Yuan et~al.(2019)Yuan, Mei, and Zhu}]{yuan2019find}
Yitian Yuan, Tao Mei, and Wenwu Zhu. 2019.
\newblock To find where you talk: Temporal sentence localization in video with
  attention based location regression.
\newblock In \emph{Proceedings of the AAAI Conference on Artificial
  Intelligence}.

\bibitem[{Zhang et~al.(2019)Zhang, Dai, Wang, Wang, and Davis}]{zhang2019man}
Da~Zhang, Xiyang Dai, Xin Wang, Yuan-Fang Wang, and Larry~S Davis. 2019.
\newblock Man: Moment alignment network for natural language moment retrieval
  via iterative graph adjustment.
\newblock In \emph{Proceedings of the IEEE/CVF Conference on Computer Vision
  and Pattern Recognition}, pages 1247--1257.

\bibitem[{Zhou et~al.(2019)Zhou, Kalantidis, Chen, Corso, and
  Rohrbach}]{zhou2019grounded}
Luowei Zhou, Yannis Kalantidis, Xinlei Chen, Jason~J Corso, and Marcus
  Rohrbach. 2019.
\newblock Grounded video description.
\newblock In \emph{Proceedings of the IEEE/CVF Conference on Computer Vision
  and Pattern Recognition}, pages 6578--6587.

\bibitem[{Zhou et~al.(2018{\natexlab{a}})Zhou, Xu, and Corso}]{zhou2018towards}
Luowei Zhou, Chenliang Xu, and Jason Corso. 2018{\natexlab{a}}.
\newblock Towards automatic learning of procedures from web instructional
  videos.
\newblock In \emph{Proceedings of the AAAI Conference on Artificial
  Intelligence}.

\bibitem[{Zhou et~al.(2018{\natexlab{b}})Zhou, Zhou, Corso, Socher, and
  Xiong}]{zhou2018end}
Luowei Zhou, Yingbo Zhou, Jason~J Corso, Richard Socher, and Caiming Xiong.
  2018{\natexlab{b}}.
\newblock End-to-end dense video captioning with masked transformer.
\newblock In \emph{Proceedings of the IEEE Conference on Computer Vision and
  Pattern Recognition}, pages 8739--8748.

\end{thebibliography}
\bibliographystyle{acl_natbib}

\clearpage
\appendix
% \pagebreak 
\section{Appendix}
\label{sec:appendix}
For ActivityNet, we tried different settings while tuning the hyper parameters. The frames are always 32 in these experiments. In the first experiment, We set the learning rate to be 0.00004 and the batch size to be 64. We update the learning rate every 150 epochs, and run 500 epochs in total. In the second experiment, we change the batch size to be 128, and update the learning rate every 200 epochs. We run 600 epochs in total. In the last two experiments, we try a different learning rate, which is 0.0004. We run the experiment for another 600 epochs with all the other parameters to be the same as in experiment 1 and 2.

% This is an appendix.

\end{document}